\documentclass[10pt,twocolumn,letterpaper]{article}

\usepackage{cvpr}
\usepackage{times}
\usepackage{epsfig}
\usepackage{graphicx}
\usepackage{amsmath}
\usepackage{amssymb}

\usepackage{url}

\cvprfinalcopy 

\def\cvprPaperID{****} 

\if\cvprfinalcopy\fi
\begin{document}

\title{Image Parsing with a Wide Range of Classes and Scene-Level Context}

\author{Marian George\\
Department of Computer Science\\
ETH Zurich, Switzerland\\
}

\maketitle
\thispagestyle{empty}

\begin{abstract}
This paper presents a nonparametric scene parsing approach that improves the overall accuracy, as well as the coverage of foreground classes in scene images. We first improve the label likelihood estimates at superpixels by merging likelihood scores from different probabilistic classifiers. This boosts the classification performance and enriches the representation of less-represented classes. Our second contribution consists of incorporating semantic context in the parsing process through global label costs. Our method does not rely on image retrieval sets but rather assigns a global likelihood estimate to each label, which is plugged into the overall energy function. We evaluate our system on two large-scale datasets, SIFTflow and LMSun. We achieve state-of-the-art performance on the SIFTflow dataset and near-record results on LMSun.

\end{abstract}
\section{Introduction}
Scene parsing is the assignment of semantic labels to each pixel in a scene image. There are various outdoor and indoor scenes (e.g., beach, highway, city street and airport) that image parsing algorithms try to label. Several systems \cite{ brostow_2008, farabet_2012,eigen_2012, grundmann_2010, Heitz_cvpr_2008,kontschieder_2014, liu_2011, munoz_2008, myeong_2012, shotton_2006, sturgess_2009, tighe_2013, yang_2014,  zhang_2010} have been designed to semantically classify each pixel in an image. Among the main challenges which face image parsing methods is that their recognition rate significantly varies among different types of classes. Background classes, which typically occupy a large proportion of the image's pixels, usually have uniform appearance and are recognised with a high rate (e.g., water, mountain, and building). Foreground classes, which typically occupy relatively few pixels in the image, have deformable shapes and can be occluded or arranged in different forms. Such classes (e.g., person, car, and sign) represent salient image regions that often capture the eye of a human observer. However, they frequently represent failure cases with recognition rates significantly lower than those of background classes.

Parametric scene parsing methods \cite{ladicky_2009, myeong_2012,sturgess_2009,Yao_2012} have achieved impressive performance on datasets with tens of labels. However, for relatively large datasets with hundreds of labels, it is more difficult to apply these methods due to expensive learning and optimisation requirements.
 
 Recently, nonparametric image parsing methods have been proposed \cite{eigen_2012, Heitz_nips_2008, singh_2013, liu_2011, tighe_2010, yang_2014} to efficiently handle the increasing number of scene categories and semantic labels. Nonparametric methods typically start by reducing the problem space from individual pixels to superpixels. First, an image set is retrieved, which contains the training images that are most visually similar to the query image. The number of candidate labels for a query image is restricted to the labels present in the retrieval set only. Second, classification likelihood scores of superpixels are obtained through visual features matching. Finally, context is enforced through minimizing an energy function which combines the data cost and knowledge about the classes co-occurences in neighboring superpixels.
 
A common challenge that faces nonparametric parsing methods is the image retrieval step. While image retrieval is useful for limiting the number of labels to consider, it is regarded as a very critical step in the pipeline \cite{singh_2013, yang_2014}. If the true labels are not included in the retrieved images, there is no chance to recover from this error. In \cite{yang_2014}, it is stated that most of the failure cases occur due to incorrect retrieval.

In this paper, we propose a novel nonparametric image parsing algorithm which targets better overall accuracy, with better recognition rates of less-represented classes. We design a system that is efficient and scalable to a continuously increasing number of labels. We make the following contributions: \begin{enumerate}
\item We improve the likelihood scores of labels at superpixels through combining classifiers. Our system combines the output probabilities of multiple classification models to produce a more balanced score for each label at each superpixel. We learn the weights for combining the scores by applying likelihood normalization method on the training set in an automatic way.
\item We incorporate semantic context in a probabilistic framework. To avoid the elimination of relevant labels that cannot be recovered at later steps, we do not construct a retrieval set. Instead, we use label costs learned from the global contextual correlation of labels in similar scenes to achieve better parsing results.
\end{enumerate}

Our system achieves state-of-the-art per-pixel recognition rates on two large-scale datasets: SIFTflow \cite{liu_2011} which contains 2688 images with 33 labels, and LMSun \cite{tighe_2010} which contains 45576 images with 232 labels.

\section{Related Work}

Several parametric and nonparametric scene parsing techniques have been proposed. Closely related to our method are the nonparametric systems which aim to achieve a wide coverage of semantic classes. The systems in \cite{tighe_2013, yang_2014, eigen_2012} adopt different techniques for boosting the overall performance of nonparametric parsing. In \cite{tighe_2013}, the authors combine region-parsing with per-exemplar SVM detector outputs. Per-exemplar detectors are used to transfer object masks into the test image for segmentation. Their system achieves impressive improvements in overall accuracy, but at the cost of expensive computational requirements. Calibrating the data terms requires batch offline training in a leave-one-out fashion, which is challenging to scale. \cite{yang_2014} and \cite{eigen_2012} explicitly add superpixels of rare classes into the retrieval set to improve their representation. The authors of \cite{yang_2014} filter the list of labels for a test image through an image retrieval step, and rare classes are enriched with more samples at query time. Our system differs in the superpixel classification technique, how we improve the recognition of rare classes, and how we apply semantic context. We promote the representation of foreground classes by merging classification costs of different contextual models, which produces more balanced label costs. We also avoid the bottleneck of image retrieval, and instead rely on global label costs in the inference step.

The usefulness of semantic context has been thoroughly explored in several visual recognition algorithms \cite{ eigen_2012, Heitz_nips_2008, Heitz_cvpr_2008, liu_2011, Rabinovich_2007,singh_2013,  yang_2014}. In the nonparametric scene parsing systems of \cite{eigen_2012, singh_2013, yang_2014}, context has been used to improve the overall labeling performance in a feedback mechanism. In \cite{eigen_2012}, initial labeling of superpixels of a query image is used to adapt the training set by conditioning on recognized background classes to improve the representation of rare classes. The goal is to improve the image retrieval set by adding back segments of \textit{rare} classes. The system in \cite{singh_2013} constructs a semantic global descriptor. Image retrieval is improved through combining the semantic descriptor with the visual descriptors. In \cite{yang_2014}, context is incorporated through building global and local context descriptors based on classification likelihood maps similar to \cite{li_2010}. Our method is different from these methods in that we do not use context at each superpixel in computing a global context descriptor, but instead we consider contextual knowledge over the image as a whole. We achieve contextually meaningful results through inferring label correlations in similar scene images. We also do not have a retrieval set which we aim to enrich. Instead, we formulate our global context in a probabilistic framework, where we compute label costs over the whole image. Also, our global context is performed online without any offline training. Another image parsing approach which does not rely on retrieval sets is \cite{Heitz_nips_2008}, where image labeling is performed by transferring annotations from a graph of patch correspondences across image sets. This approach, however, requires large memory which is difficult to scale for large datasets like SIFTflow and LMSun.

Our approach is inspired from combining classifiers techniques \cite{kittler_1998} in machine learning, which have been shown to boost  the strengths of single classifiers. Several fusion techniques have been successfully used in different areas of computer vision, like face detection \cite{viola_2004}, multi-label image annotation \cite{qi_2007}, object tracking \cite{Zhaozheng_2008}, and character recognition \cite{Ho_1994}. However, the constituent classifiers and the mechanisms for combining them are quite different from our framework and the other techniques are only demonstrated on small datasets.


\section{Baseline Parsing Pipeline} \label{sec:approach}
In this section, we present an overview of our baseline image parsing system, which consists of three steps: feature extraction (sec. \ref{sec:features}), label likelihood estimation at superpixels (sec. \ref{sec:classifier}), and inference (sec. \ref{sec:mrf}).

Following that, we present our contributions of improving likelihoods at superpixels and computing label costs for scene-level global context in sections \ref{sec:combining_classifiers} and \ref{sec:label_costs} respectively.

\subsection{Segmentation and Feature Extraction} \label{sec:features}
To reduce the problem space, we divide the image into superpixels.
We start by extracting superpixels from images using the efficient graph-based method of \cite{Felzenszwalb_2004}. For each superpixel, we extract 20 types of local features to describe its shape, appearance, texture, color, and location, following the method of \cite{tighe_2010}. In addition to these features, we extract Fisher Vector (FV) \cite{Perronnin_2010} descriptors at each superpixel using the VLFeat library \cite{vlfeat}. We compute 128-dimensional dense SIFT feature descriptors on 5 patch sizes (8, 12, 16, 24, 30). We build a dictionary of size 1024 words. We then extract the FV descriptors and apply PCA to reduce their size to $512$ dimensions. Each superpixel is described by a 2202-dimensional feature vector.

\begin{figure*} [t]
\begin{center}
\includegraphics[width=0.9\textwidth]{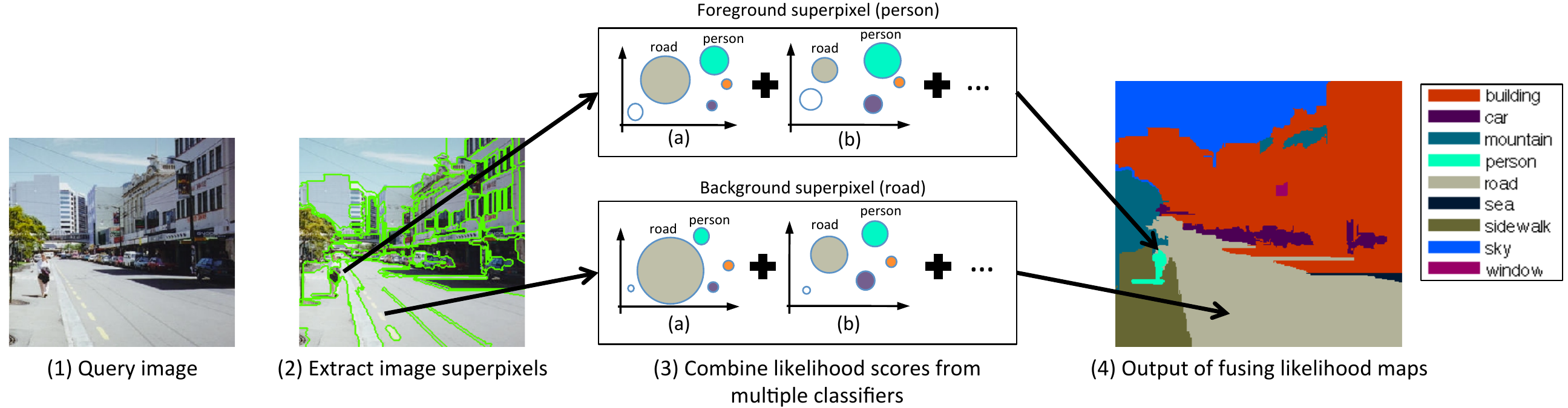}
\end{center}
 \caption{Overview of the fusing classifiers approach. Likelihood scores from multiple models (3a) and (3b) are combined to produce the final likelihoods at superpixels. Likelihood scores of foreground classes (e.g. person) are boosted via our combination technique. The unbalanced (skewed) model in (3a)  produces biased likelihoods towards background classes (e.g. road). This is reflected in the much larger score (bigger circle) for the road class when compared to the person class and other less-represented classes. For the balanced classifier in (3b), the scores are more balanced and less-represented classes get a higher chance (bigger circle) of being recognized.}
\label{fig:system_overview}
\end{figure*}

\subsection{Label Likelihood Estimation} \label{sec:classifier}
We use the extracted features at the previous step to compute label likelihoods at each superpixel.
Different from traditional methods, we do not restrict the potential labels for a test image. We instead compute the likelihood data term for each class label $c \in C$, where $C$ is the total number of classes in the dataset. The normalized cost $D(l_{s_i}=c|s_i)$ of assigning label $c$ to superpixel $s_i$ is given by:
\begin{equation}
D(l_{s_i}=c|s_i) = 1 - \frac{1}{1 + e^{-L_{unbal}(s_i,c)}}, 
\end{equation}
where $L_{unbal}(s_i,c)$ is the log-likelihood ratio score of label $c$, given by $L_{unbal}(s_i,c) = \frac{1}{2} log (P(s_i|c) / P(s_i|\bar{c}))$, where $\bar{c} = C \medspace \backslash \medspace c$ is the set of all labels except $c$, and $P(s_i|c)$ is the likelihood of superpixel $s_i$ given $c$. We learn a boosted decision tree (BDT) \cite{Collins_2002} model to obtain the label likelihoods $L_{unbal}(s_i,c)$. For implementation, we use the publicly available boostDT \footnote{http://web.engr.illinois.edu/~dhoiem/software/} library. At this stage, we train the BDT model using all superpixels in the training set, which represent an unbalanced distribution of class labels $C$. 

\subsection{Smoothing and Inference} \label{sec:mrf}

We formulate our optimization problem as that of maximum a posteriori (MAP) estimation  of the final labeling $L$ using Markov Random Field (MRF) inference. Using only the estimated likelihoods in the previous section to classify superpixels yields noisy classifications. Adding a smoothing term $V(l_{s_i},l_{s_j})$ to the MRF energy function attempts to overcome that issue by punishing neighboring superpixels having semantically irrelevant labels. 
Our baseline attempts to minimize the following energy function:
\begin{equation} \label{equ:mrf}
E(L) = \sum_{s_i \in S}{D(l_{s_i}=c|s_i) } + \lambda \sum_{(i,j)\in A}{V(l_{s_i},l_{s_j})}.
\end{equation}
where $A$ is the set of adjacent superpixel indices and $V(l_{s_i},l_{s_j})$ is the penalty of assigning labels $l_{s_i}$ and $l_{s_j}$ to two neighboring pixels, computed from counts in the training set combined with the constant Potts model following the approach of \cite{tighe_2010}. $\lambda$ is the smoothing constant. We perform inference using the $\alpha$-expansion method with the code of \cite{boykov_2001,Kolmogorov_2004,Boykov_2004}.

In the next two sections, we present our main contributions of how we improve the superpixel classification step (section \ref{sec:combining_classifiers}) and how we incorporate scene-level context to achieve better results (section \ref{sec:label_costs}).

\section{Improving Superpixel Label Costs} \label{sec:combining_classifiers}
While foreground objects are usually the most noticeable regions in a  scene image, they are often misclassified by parsing algorithms. For example, in a city street scene, a human viewer would typically first notice the people, signs and cars before noticing the buildings and road. However, for scene parsing algorithms, foreground regions are often misclassified as being part of the surrounding background due to two main reasons. First, in the superpixel classification step, any classifier would naturally favor more dominant classes to minimize the overall training error. Second, in the MRF smoothing step, many of the superpixels which were correctly classified as foreground objects, are smoothed out by neighboring background pixels.

We propose to improve the label likelihood score at each superpixel to achieve a more accurate parsing output. We design different classifiers that offer complementary information about the data. All the designed models are then combined to derive a consensus decision. The overview of our fusing classifiers approach is shown in Figure \ref{fig:system_overview}. At test time, the label likelihood scores of all the BDT models are merged to produce the final scores at superpixels. 
 
\begin{figure} [h]
\begin{center}
	\includegraphics[width= 0.45\textwidth]{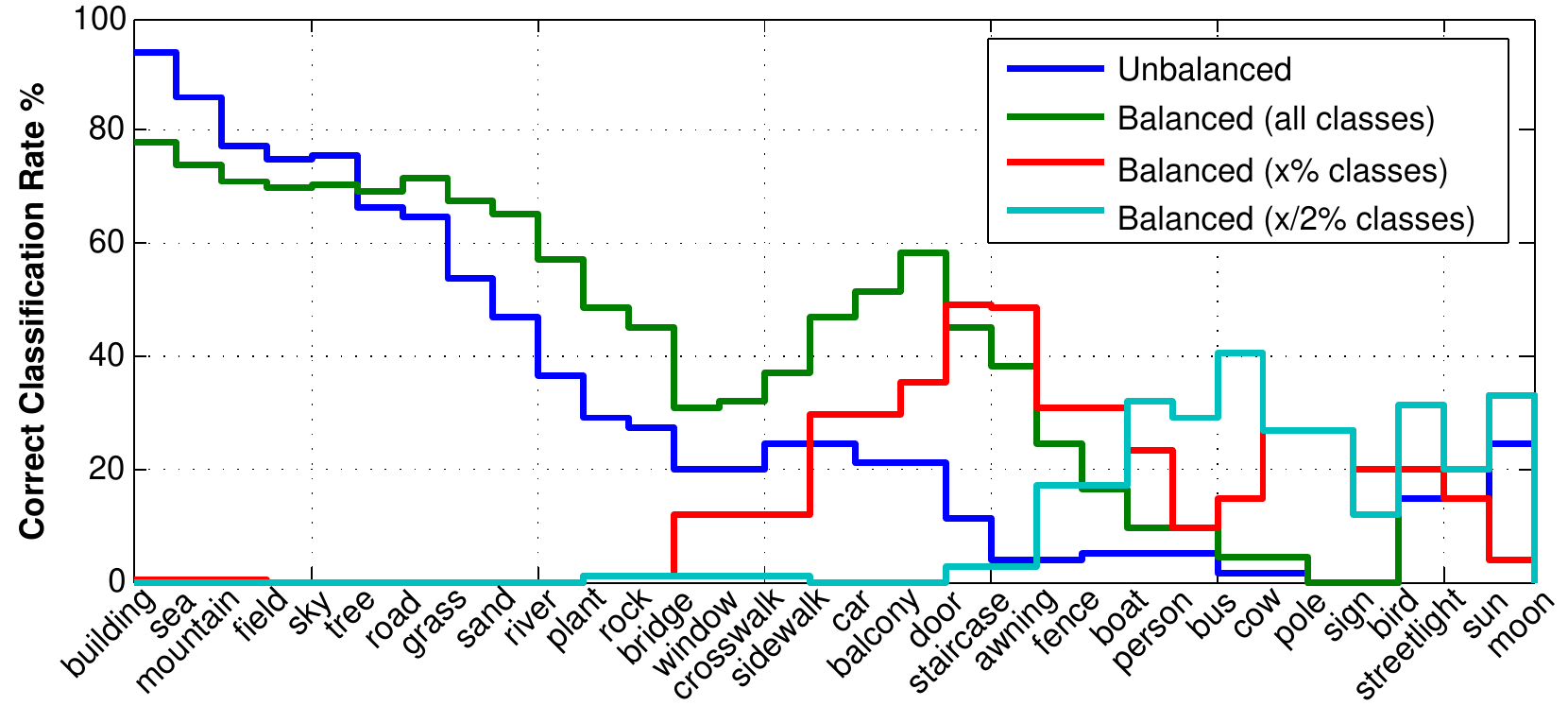}
\end{center}
 \caption{Classification rates (\%) of individual classes for the different classification models trained on SIFTflow. Classes are ordered in descending order by the mean number of pixels they occupy (frequency) in scene images. Our goal is to decrease the correlation between the trained models.}
\label{fig:correlation}
\end{figure}

\subsection{Fusing Classifiers} \label{sec:fusing_classifiers}
Our method is inspired from ensemble classifier techniques that train multiple classifiers and combine them to reach a better decision. Such techniques are specifically useful if the classifiers are different \cite{kittler_1998}. In other words, the error reduction is related to the uncorrelation between the trained models \cite{tumer_1996}, i.e. the overall error is reduced if the classifiers misclassify different data points. Also, it has been shown that partitioning the training set performs better than partitioning the feature space for large datasets \cite{tumer_1996}.

We have observed that the classification error of a given class is related to the mean number of pixels it occupies in the scene images, as shown by the blue line in Figure \ref{fig:correlation}. This agrees with the findings of previous methods \cite{tighe_2013, yang_2014} that the classification error rate is related to the frequency of classes in the training set. However, we go beyond that by considering the frequency of the classes on the image level, which targets the problem of smoothing out the less-represented classes by a neighbouring background class.

To this end, we train three BDT models with the following training data criteria: (1) a balanced subsample of all classes $C$ in the dataset, (2) a balanced subsample of classes occupying an average of less than $x\%$ of their images, and (3)  a balanced subsample of classes occupying an average of less than $\left \lceil{x/2}\right \rceil \%$ of their images.

The motivation beyond these choices is to reduce the correlation between the trained BDT models as shown in Figure \ref{fig:correlation}. While the unbalanced classifier mainly misclassifies the less-represented classes, the balanced classifiers recover some of these classes while making more mistakes on the more represented classes. By combining the likelihoods from all the classifiers, a better overall decision is reached that improves the overall coverage of classes (Figure \ref{fig:system_overview}). We observed that the addition of more classifiers did not improve the performance for any of our datasets. 

The final cost of assigning a label $c$ to a superpixel $s_i$ can then be represented as the combination of the likelihood scores of all classifiers:
\begin{equation}
D(l_{s_i}=c|s_i) = 1 - \frac{1}{1 + e^{-L_{comb}(s_i,c)}}
\end{equation}
where $L_{comb}(s_i,c)$ is the combined likelihood score obtained by the weighted sum of the scores from all classifiers:
\begin{equation}
L_{comb}(s_i,c) = \sum_{j=1,2,3,4}{w_{j}(c) \thinspace L_{j}(s_i,c)},
\end{equation}
where $L_{j}(s_i,c)$ is the score from the $j^{th}$ classifier, and $w_j(c)$ is the normalized weight of the likelihood score of class $c$ in the $j^{th}$ classifier.

\subsection{Normalized Weight Learning} \label{sec:weight_learning}
We learn the weights $\textbf{w}\equiv[w_j(c)]$ of all classes $C$ in offline settings using the training set. We compute the weights separately for each classifier. The weight $\tilde{w}_{j}(c)$ of class $c$ for the $j^{th}$ classifier is computed as the average ratio of the sum of all likelihoods of class $c$, to the sum of all likelihoods of all classes $c_i \in C\backslash c$ of all superpixels $s_i \in S$:
\begin{equation}
\tilde{w}_{j}(c) = \frac{|C_{j}|}{C}\frac{\sum_{s_i \in S}{L_{j}(s_i,c)}}{\sum_{s_i \in S}{\sum_{c_i \in C\backslash c}{L_{j}(s_i,c_i)}}}
\end{equation}
where $|C_{j}|$ is the number of classes covered by the $j^{th}$ classifier and not covered by any other classifier with a smaller number of classes.

The normalized weight \textbf{$w_j(c)$} of class $c$ can then be computed as:  $w_j(c) =\tilde{w}_{j}(c) / \sum_{j=1,2,3,4}{\tilde{w}_{j}(c)}$. Normalizing the output likelihoods in this manner gives a better chance for all classifiers to be considered in the result with an emphasis on less-represented classes. In sec. \ref{sec:experiments}, we show the superior performance of our fusion scheme to other traditional fusion mechanisms: averaging and median rule.
\section{Scene-Level Global Context} \label{sec:label_costs}
When exploiting scene parsing problems, it is useful to incorporate the semantics of the scene in the labeling pipeline. For example, if we know that a given scene is a beach scene, we will expect to find labels like sea, sand, and sky with a much higher probability than expecting to find labels like car, building, or fence.
We use the initial labeling results of a test image in estimating the likelihoods of all labels $c \in C$ (sec. \ref{sec:global_costs_1}). The likelihoods are estimated globally over an image, i.e. there is a unique cost per label per image. We then plug the global label costs into a second MRF inference step to produce better results (sec. \ref{sec:global_costs_2}).

Our approach, unlike previous methods, does not limit the number of labels to those present in the retrieval set but instead uses the set to compute the likelihood of class labels in a k-nn fashion. The likelihoods are normalized by counts over the whole dataset and smoothed to give a chance to labels not in the retrieval set. We also employ the likelihoods in MRF optimization, not for filtering the number of labels.

\begin{figure} [h]
\begin{center}
	\includegraphics[width=0.45\textwidth]{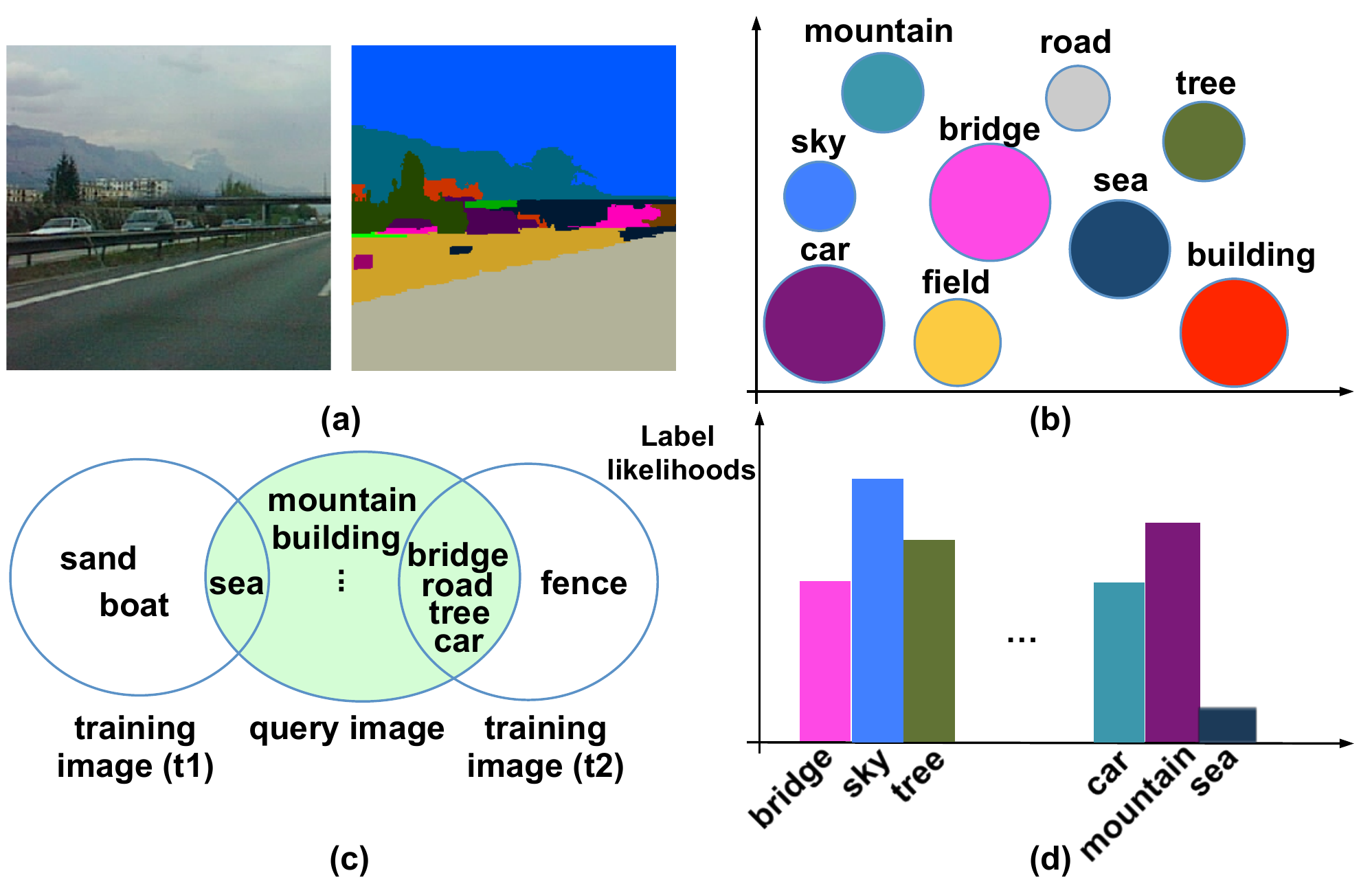}
\end{center}
 \caption{Scene-level global context. (a) The initial labeling of a query image is used to (b) assign weights to the unique classes in the image. A class with a bigger weight is represented by a larger circle. (c) Training images are ranked by the \textit{weighted} size of intersection of their class labels with the query. (d) Global label likelihoods are computed through label counts in the top-ranked images.}
\label{fig:global_label_cots}
\end{figure}

\subsection{Context-Aware Global Label Costs} \label{sec:global_costs_1}
We propose to incorporate semantic context through using label statistics instead of global visual features. The intuition behind such choice is that ranking by global visual features often fails to retrieve similar images on the scene level \cite{tighe_2010, yang_2014}. For example, a highway scene could be confused with a beach scene with road pixels misclassified as sand. However, ranking by label statistics, given a relatively good initial labeling, retrieves more semantically similar images that aim to remove outlier labels (e.g., sea pixels in street scene), and recover missing labels in a scene.

For a given test image $I$, minimizing the energy function in equation \ref{equ:mrf} produces an initial labeling $L$ of the superpixels in the image. If $C$ is the total number of classes in the dataset, let $T \subset C$ be the set of unique labels which appear in $L$, i.e. $T = \{t \medspace | \exists s_i: l_{s_i} =t\}$, where $s_i$ is a superpixel with index $i$ in the test image, and $l_{s_i}$ is the label of $s_i$. We exploit semantic context in a probabilistic framework, where we model the conditional distribution $P(c|T)$ over class labeling $C$ given the initial global labeling of an image $T$. We compute $P(c|T) \forall c \in C$ in a $K$-nn fashion:
\begin{equation} \label{equ:global_cost}
P(c|T) = \frac{(1+n(c,K_T))/n(c,S)}{(1+n(\bar{c},K_T))/|S|},
\end{equation}
where $K_T$ is the $K$-neighborhood of initial labeling $T$, $n(c,X)$ is the number of superpixels with label $c$ in $X$, $n(\bar{c}, X)$ is the number of superpixels with all labels except $c$ in $X$, and $|S|$ is the total number of superpixels in the training set. We normalize the likelihoods and  add a smoothing constant of value $1$.

To get the neighborhood $K_T$, we rank the training images by their distance to the query image. The distance between two images is computed as the \textit{weighted} size of intersection of their  class labels, intuitively reflecting that the neighbors of $T$ are images with many shared labels with those in $T$. We assign a different weight to each class in $T$ in such a way to favor less-represented classes.

As shown in Figure \ref{fig:global_label_cots}, our algorithm works in three steps. It starts by (1) assigning a weight $\omega_{t}$ to each class $t \in T$, which is inversely proportional to the number of superpixels in the test image with label $t$: $\omega_{t} = 1-\frac{n(t,I)}{|I|}$, where $n(t,I)$ is the number of superpixels in the test image with label $l_{s_i} = t$, and $|I|$ is the total number of superpixels in the image. 
Then, (2) training images are ranked by the weighted size of intersection of their class labels with the test image. Finally, (3) the global label likelihood $L_{global}(c) = P(c|T)$ of each label $c \in C$ is computed using equation \ref{equ:global_cost}.


Computing the label costs is done online for a query image without any batch offline training. Our method improves the overall accuracy by using only the ground truth labels of training images without any global visual features.

\subsection{Inference with Label Costs} \label{sec:global_costs_2}
Once we obtained the likelihoods $L_{global}(c)$ of each class $c \in C$, we can define a label cost $H(c) = -log(L_{global}(c))$. Our final energy function becomes:
\small
\begin{equation} \label{equ:new_mrf}
E(L) = \sum_{s_i \in S}{D(l_{s_i}=c|s_i) }+\lambda \sum_{(i,j)\in A}{V(l_{s_i},l_{s_j})} + \sum_{c \in C}{H(c).\delta{(c)}},
\end{equation}
\normalsize
where $\delta{(c)}$ is the indicator function of label $c$: 
\begin{equation} \nonumber
\delta{(c)} = \left\{
\begin{array}{l l}
1 & \quad \exists s_i:l_{s_i} = c \\
0 & \quad otherwise
\end{array} \right.
\end{equation}
We solve equation \ref{equ:new_mrf} using $\alpha$-expansion with the extension method of \cite{Delong_2012} to optimize label costs. Optimizing the energy function in equation \ref{equ:new_mrf} effectively minimizes the number of unique labels in a test image to those which have low label costs, i.e. which are most relevant to the scene.
\section{ Experiments} \label{sec:experiments}
We ran our experiments on two large-scale datasets: SIFTflow \cite{liu_2011} and LMSun \cite{tighe_2010}. SIFTflow has 2,488 training images and 200 test images. All images are of outdoor scenes of size 256x256 with 33 labels. LMSun contains both indoor and outdoor scenes, with a total of 45,676 training images and 500 test images. Image sizes vary from 256x256 to 800x600 pixels with 232 labels.

We use the same evaluation metrics and train/test splits as previous methods. We report the per-pixel accuracy (the percentage of pixels of test images that were correctly labeled), and per-class recognition rate (the average of per-pixel accuracies of all classes). We evaluate the following variants of our system: (i) \textit{baseline}, as described in sec. \ref{sec:approach}, (ii) \textit{baseline (with balanced BDT)}, which is the baseline approach using a balanced classifier, (iii) \textit{baseline + FC (NL fusion)}, which is the baseline in addition to the fusing classifiers with normalized-likelihood (NL) weights in sec. \ref{sec:combining_classifiers}, and (iv) \textit{full}, which is baseline + fusing classifiers + global costs. To show the effectiveness of our fusion method (sec. \ref{sec:weight_learning}), we report the results of (v) \textit{baseline + FC (average fusion)}, which is fusing classifiers by averaging their likelihoods, and (vi) \textit{baseline + FC (median fusion)}, which is fusing classifiers by taking the median of their likelihoods. We also report results of (vii) \textit{full (without FV)}, which is full system without using the Fisher Vector features.

We fix $x=5$ (sec.\ref{sec:fusing_classifiers}), a value that was obtained through empirical evaluation on a small subset of the training set.
\subsection{Results}
\begin{figure*} [t!]
\begin{center}
	\includegraphics[width=0.9\textwidth]{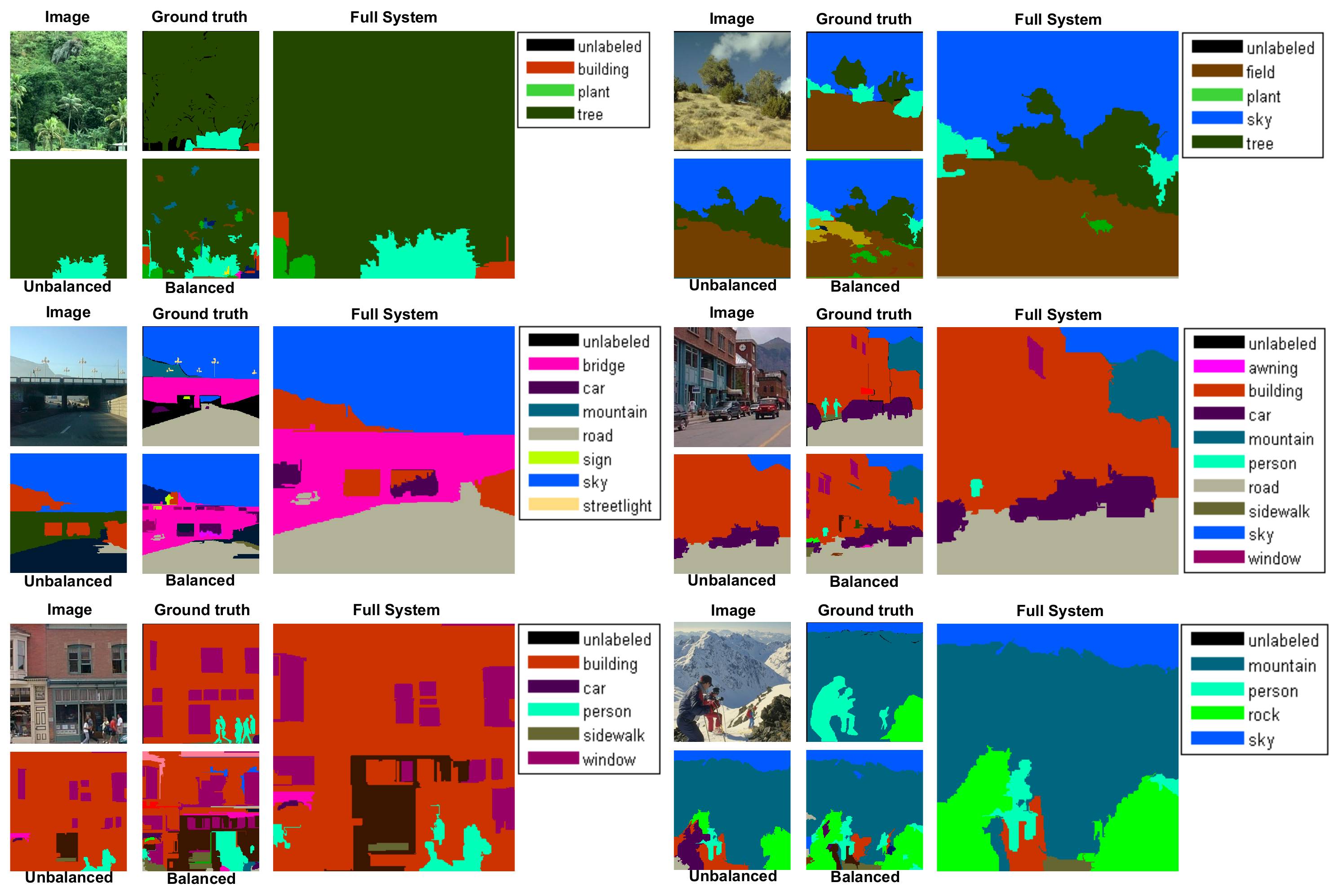}
\end{center}
 \caption{Examples of parsing results on the SIFTflow dataset (best viewed in color). Top left is the original image, on its right is the ground truth labeling, bottom left is the output from the baseline, on its right the output of the balanced classifier. Finally, the output of the full system is on the far right (third column). The unbalanced classifier often misses the foreground classes by oversmoothing the results. The balanced classifier performs better with foreground classes, but yields more noisy classification. The full system combines the benefits of both classifiers, improving both the overall accuracy and the coverage of foreground classes (e.g., building, bridge, window, and person)}
\label{fig:sift_flow_samples}
\end{figure*}

We compare our results with state-of-the-art methods on SIFTflow in Table \ref{table:sift_flow}. We have set $K=64$ top-ranked training images for computing the  global context likelihoods (sec. \ref{sec:global_costs_1}). Our full system achieves 81.7\% per-pixel accuracy, and 50.1\% per-class accuracy, which outperforms the state-of-the-art method of \cite{yang_2014} (79.8\% / 48.7\%). Results show that our fusing classifiers step significantly boosts the coverage of foreground classes, where the per-class accuracy increases by around 15\% over the baseline method. Our semantic context (sec. \ref{sec:label_costs}) improves both the per-pixel and per-class accuracies through optimizing for fewer labels which are more semantically meaningful. Fisher Vectors improved the recognition by around $3\%$. In Figure \ref{fig:sift_flow_samples}, we show examples of parsing results on the SIFTflow dataset. 
\begin{table} [h]
\begin{center}
\footnotesize
\begin{tabular}{|l|c|c|}
\hline
Method & Per-pixel & Per-class \\
\hline\hline
Liu et al. \cite{liu_2011} & 76.7 & N/A\\
Farabet et al. \cite{farabet_2012} & 78.5 & 29.5 \\
Farabet et al. \cite{farabet_2012} balanced & 74.2 & 46.0\\
Eigen and Fergus \cite{eigen_2012} & 77.1 & 32.5 \\
Singh and Kosecka \cite{singh_2013} & 79.2 & 33.8 \\
Tighe and Lazebnick \cite{tighe_2010} & 77.0 & 30.1 \\
Tighe and Lazebnick \cite{tighe_2013} & 78.6 & 39.2 \\
Yang et al. \cite{yang_2014} & 79.8 & 48.7 \\
\hline
Baseline & 78.3 & 33.2 \\
Baseline (with balanced BDT) & 76.2 & 45.5\\
Baseline + FC (NL fusion) & 80.5 & 48.2 \\
Baseline + FC (average fusion) & 78.6 & 46.3 \\
Baseline + FC (median fusion) & 77.3 & 46.8 \\
Full without Fisher Vectors & 77.5 & 47.0 \\
\hline
\textbf{Full} & \textbf{81.7} &  \textbf{50.1} \\
\hline
\end{tabular}
\end{center}
\caption{Comparison with state-of-the-art per-pixel and per-class accuracies (\%) on the SIFTflow dataset.}
 \label{table:sift_flow}
\end{table}

\begin{figure} [h]
\begin{center}
	\includegraphics[width= 0.42\textwidth]{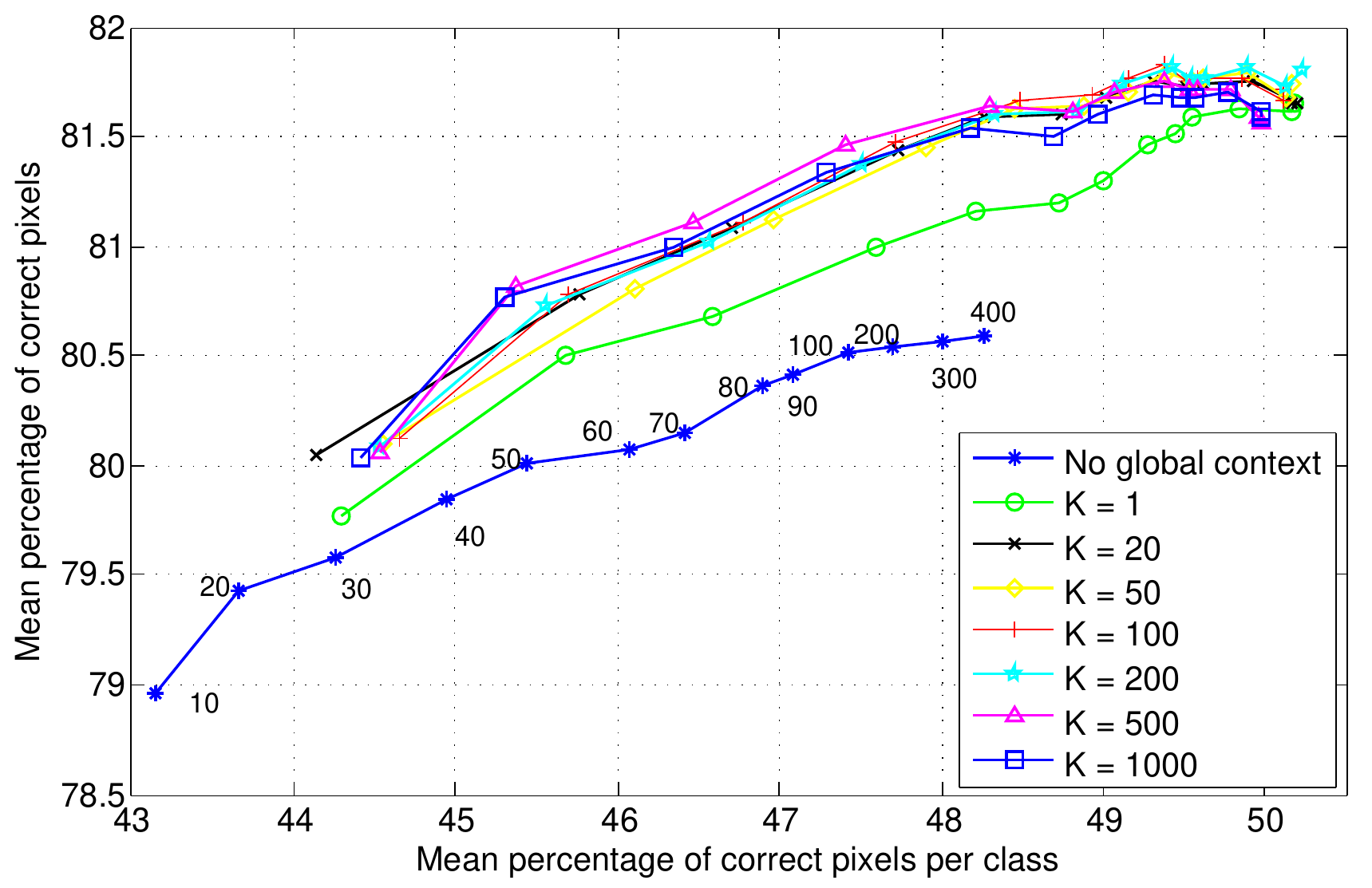}
\end{center}
 \caption{Analysis of the performance when varying the number of trees for training the BDT model, at different values of top $K$ images for the global context step on the SIFTflow dataset. The y-axis shows the per-pixel accuracies (\%) and the x-axis show the per-class accuracies (\%) for different numbers of trees.}
\label{fig:global_analysis}
\end{figure}

\begin{figure*} [t] 
\begin{center}
	\includegraphics[width=0.9\textwidth]{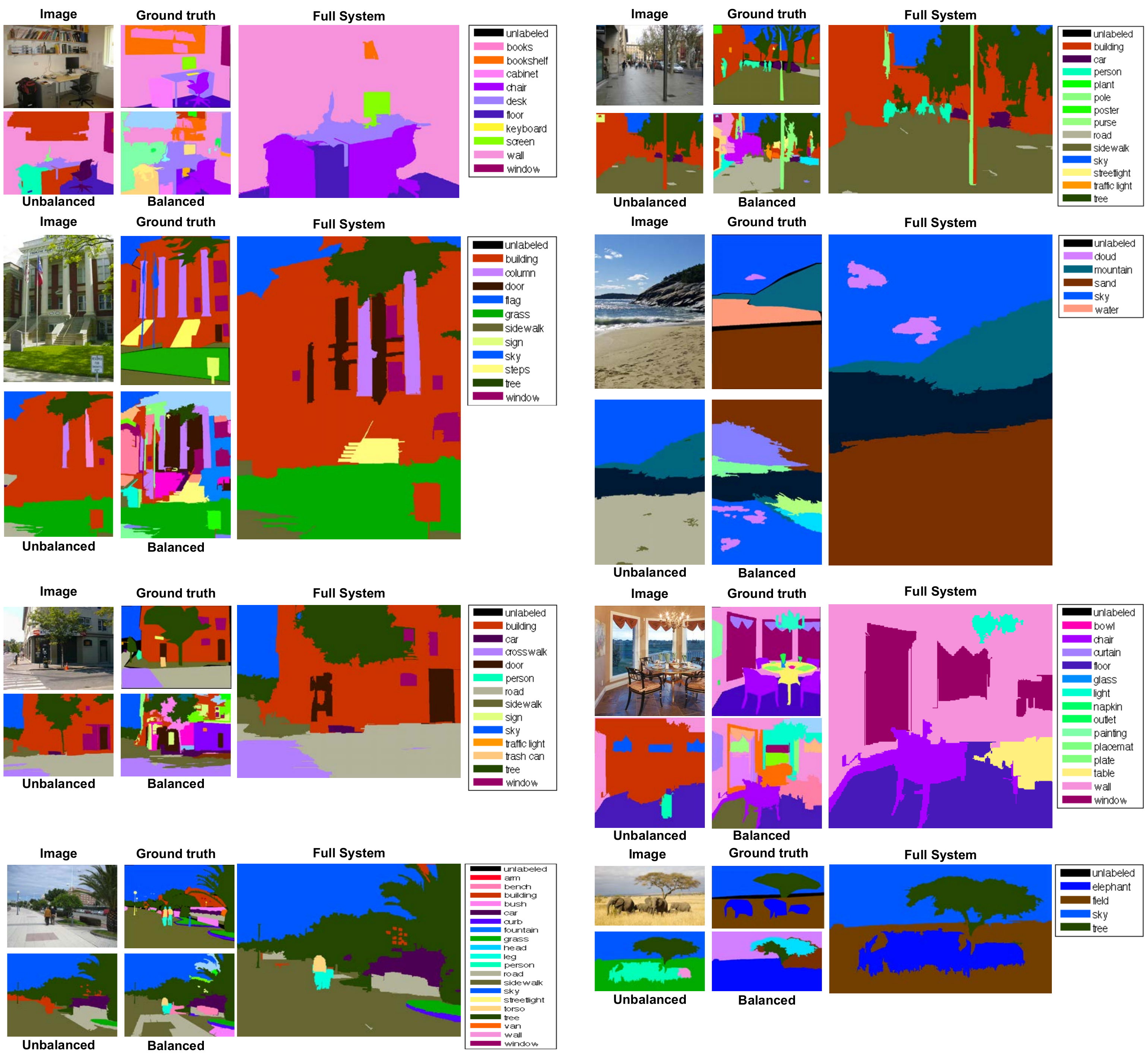}
\end{center}
 \caption{Examples of parsing results on the LMSun dataset (best viewed in color). The layout of the results is the same as in Fig. \ref{fig:sift_flow_samples}. Foreground classes (e.g. screen, sidewalk, person, torso, pole, cloud, table, light, and elephant) are successfully recognized by our system.}
\label{fig:lmsun_samples}
\end{figure*}

Table \ref{table:LMSun} compares the performance of the same variants of our system with the state-of-the-art methods on the large-scale LMSun dataset. LMSun is more challenging than SIFTflow in terms of the number of images, the number of classes, and the presence of both indoor and outdoor scenes. Accordingly, we use a larger value of $K=200$ in equation \ref{equ:global_cost}. Our method achieves near record performance in per-pixel accuracy ($61.2\%$), while placing second in per-class accuracy. The effectiveness of the fusing classifiers technique is shown in the improvement of both per-pixel (by $3\%$) and per-class (by $4.5\%$) accuracies over the baseline system. The global context step improves the class coverage by around $2\%$. Figure \ref{fig:lmsun_samples} shows the output of our scene parsing system on some images from LMSun.

\begin{table} [h]
\begin{center}
\footnotesize
\begin{tabular}{|l|c|c|}
\hline
Method & Per-pixel & Per-class \\
\hline\hline
Tighe and Lazebnick \cite{tighe_2010} & 54.9 & 7.1 \\
Tighe and Lazebnick \cite{tighe_2013} & \textbf{61.4} & 15.2 \\
Yang et al. \cite{yang_2014} & 60.6 & \textbf{18.0} \\
\hline
Baseline & 57.3 & 9.5 \\
Baseline (with balanced BDT) & 45.4 & 13.8\\
Baseline + FC (NL fusion) & 60.0 & 14.2 \\
Baseline + FC (average fusion) & 60.5 & 11.4 \\
Baseline + FC (median fusion) & 59.2 & 14.7 \\
Full without Fisher Vectors & 58.2 & 13.6 \\
\hline
\textbf{Full} & 61.2 & 16.0 \\
\hline
\end{tabular}
\end{center}
\caption{Comparison with state-of-the-art per-pixel and per-class accuracies (\%) on the LMSun dataset.}
\label{table:LMSun}
\end{table}

\begin{figure} [h]
\begin{center}
	\includegraphics[width= 0.47\textwidth]{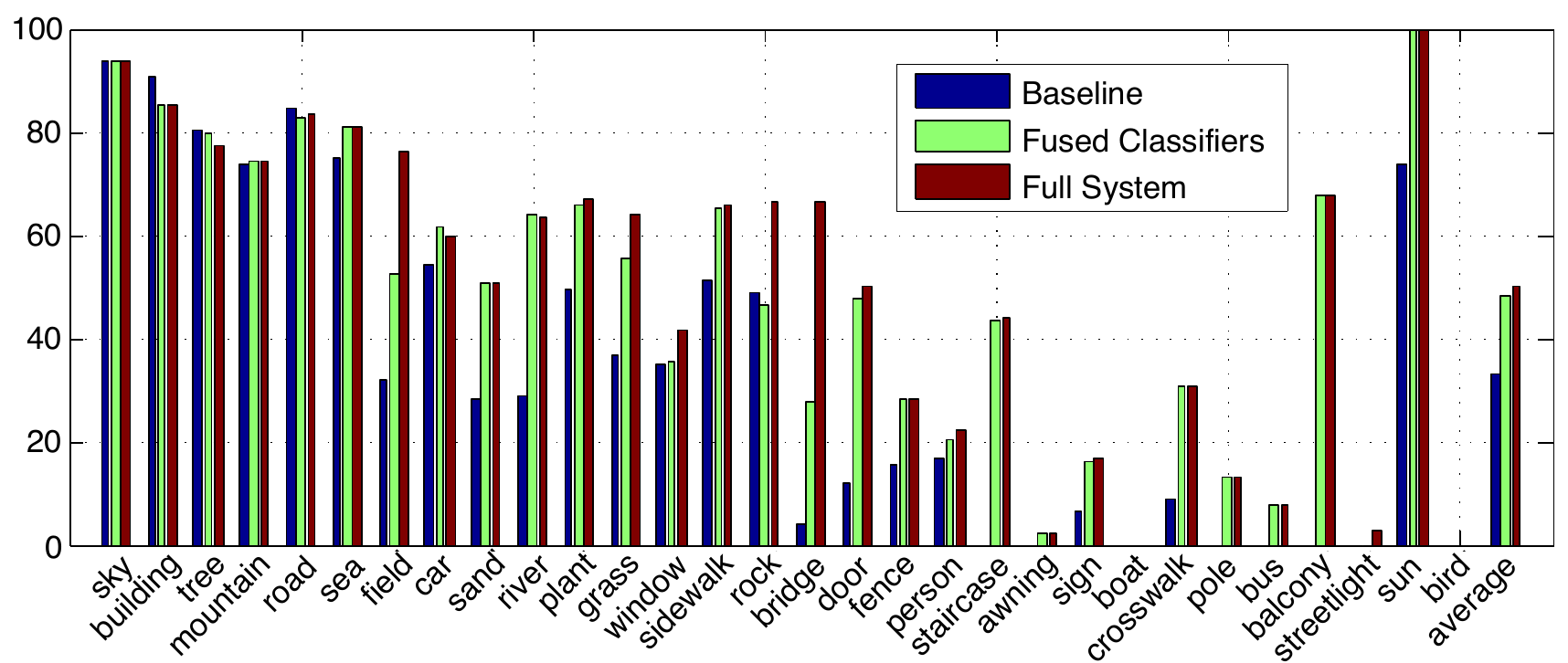}
\end{center}
 \caption{Classification rates (\%) of individual classes for the baseline, fused classifiers, and the full system on SIFTflow. Classes are sorted from most frequent to least frequent. }
\label{fig:heatmap}
\end{figure}

We next analyze the performance of our system when varying the number of trees $T$ for training the BDT model (sec. \ref{sec:fusing_classifiers}), and the number of top training images $K$ in the global label costs (sec. \ref{sec:global_costs_1}). Figure \ref{fig:global_analysis} shows the per-pixel accuracy (on the y-axis) and the per-class accuracy (on the x-axis) as a function of $T$ for a variety of $K$s. Increasing the value of $T$ generally produces better classification models that better describe the training data. At $T \geq 400$, performance levels off. 
As shown, our global label costs consistently improve the performance over the baseline method with no global context. Using more training images (higher $K$) improves the performance through considering more semantically-relevant scene images. However, performance starts to decrease for very high values of $K$ (e.g., $K=1000$) as more noisy images start to be added.

Figure \ref{fig:heatmap} shows the per-class recognition rate for the baseline, combined classifiers, and the full system on SIFTflow. Our fusing classifiers technique produces more balanced likelihood scores that cover a wider range of classes. The semantic context step removes outlier labels and recovers missing labels, which improves the recognition rates of both common and rare classes. Recovered classes include field, grass, bridge, and sign. Failure cases include extremely rare classes, e.g. cow, bird, desert, and moon.

\subsection{Running Time}
We analyzed the runtime performance for both SIFTflow and LMSun (without feature extraction) on a four-core 2.84GHz CPU with 32GB of RAM without code optimization. For the SIFTflow dataset, training the classifier takes an average of 15 minutes per class. We run the training process in parallel. The training time highly depends on the feature dimensionality. At test time, superpixel classification is efficient, with an average of 1 second per image. Computing global label costs takes 3 seconds. Finally, MRF inference takes less than one second. We run MRF inference twice for the full pipeline. LMSun is much larger than SIFTflow. It takes 3 hours for training the classifier, less than a minute for superpixel classification per image, less than 1 minute for MRF inference, and $\sim$2 minutes for global label cost computation.

\subsection{Discussion}
Our scene parsing method is generally scalable as it does not require any offline training in a batch fashion. However, the time required for training a BDT classifier increases linearly with increasing the number of data points. This is challenging with large datasets like LMSun. Randomly subsampling the dataset  has a negative impact on the overall precision of the classification results. We plan to investigate alternative approaches like \cite{singh_2012} of mining discriminative data points that better describe each class. Our system still faces challenges in trying to recognize very less-represented classes in the dataset (e.g., bird, cow, and moon). This could be handled via better contextual models per query image. 

\section{Conclusion} \label{sec:conclusion}
In this work, we have presented a novel scene parsing algorithm that improves the overall labeling accuracy, without smoothing away foreground classes which are important for human observers. Through combining likelihood scores from different classification models, we have successfully boosted the strengths of individual models, thus improving both the per-pixel, as well as the per-class accuracies. To avoid eliminating correct labels through image retrieval, we have encoded global context into the parsing process in a probabilistic framework. We have extended the energy function to include global label costs that achieve more semantically meaningful parsing output. Experiments have shown the superior performance of our system on the SIFTflow dataset and comparable performance to state-of-the-art methods on the LMSun dataset.

\section*{Acknowledgements} We thank Friedemann Mattern and Christian Floerkemeier for the useful discussions, and the CVPR reviewers for their insightful feedback.

{\small
\bibliographystyle{ieee}
\bibliography{egbib}
}

\end{document}